\documentclass[sigconf,nonacm]{acmart}

\AtBeginDocument{}

\usepackage{xcolor}
\usepackage{enumitem}
\usepackage{multirow}
\usepackage{array}
\usepackage{listings}
\begin{document}

\title{Measuring Where Intent Gets Lost in Agentic Smart Contract Synthesis}

\author{Abhinav Goel}
\affiliation{\institution{Columbia University}\city{New York}\state{NY}\country{USA}}
\author{Agostino Capponi}
\affiliation{\institution{Columbia University}\city{New York}\state{NY}\country{USA}}
\author{Alfio Gliozzo}
\affiliation{\institution{IBM T.J. Watson Research Center}\city{Yorktown Heights}\state{NY}\country{USA}}
\author{Chaitya Shah}
\affiliation{\institution{Columbia University}\city{New York}\state{NY}\country{USA}}
\renewcommand{\shortauthors}{Goel et al.}

\begin{abstract}
LLM agents can turn a plain-English description into code that compiles, but whether the code behaves as required is rarely measured with validated instruments. We study this for smart contracts, the programs behind Web3 applications that move funds and cannot be changed once deployed. Starting from 600 existing contracts, we have an LLM describe each and measure implementation fidelity, the share of replayed transactions on which a generated contract behaves like the original. Three findings concern measurement. LLM judges, static analysis and LLM-based testers fail on planted mistakes, an answer key written from function names alone catches most conventional ones, and scoring only the functions all systems share, or adapting calls to each contract, hides omissions. Under exact replay, information supplied before generation matters most. Answers from a simulated client that reads the original contract raise fidelity by 19 to 23 points for three generator families, an optimistic estimate of what real authors would provide, and three to five well-chosen questions recover 60 to 75\% of that gain. Stages after generation add at most about one point, and an automated security-repair loop costs 12.8 points. An elicitation-first pipeline reaches 72.9\% on 500 held-out contracts, against 53.9\% for one-shot generation and 39.3\% for the original pipeline. We release the harness, tasks and scores.
\end{abstract}


\maketitle

%% ===================================================================
\section{Introduction}
%% ===================================================================

Agentic pipelines that turn natural-language requirements into code are now routine, and for many tasks the code compiles and runs. Whether it does what its author meant is a different question, and it is the one that matters when the code moves money. Smart contracts make this concrete. They are the back end of Web3 applications that people use through browsers and wallets, they hold and transfer value autonomously, cannot be patched after deployment, and fail expensively~\cite{atzei2017survey, chen2020survey, luu2016making}. Several systems now synthesize Solidity from requirements with multi-agent pipelines~\cite{luo2025fsmscg, chen2026solagent, zhang2026lever, priorpipeline2026}. Agents that act on Web services for a user face a related problem, since a natural-language request also underdetermines the intended behavior; contracts are a convenient place to study it because every action can be replayed exactly in a sandbox.

Evaluating such systems is hard for three reasons. The only statement of what is wanted is usually a description written in English, with no implementation to compare against. The instruments commonly used as proxies, such as LLM judges, static analyzers and compilation, measure something other than intent. And much of what a contract must do follows universal conventions that a model knows without reading the description, so an evaluation can reward conventions and look like it rewards understanding.

This paper is an evaluation study. We start from our earlier seven-phase pipeline~\cite{priorpipeline2026}, whose evaluation used an LLM rubric judge and an LLM security gate~\cite{priorpipeline2026, priorbenchmark2026}, ask how its output should be measured, and use the resulting instruments to locate where information is lost and which stages recover it. We distinguish author intent, compliance with a written specification and implementation fidelity, agreement with a given implementation on the same transactions. Because our tasks start from existing contracts, implementation fidelity is the construct we measure; it reflects intent only where the original contract embodies what its author wanted. Our central lesson is methodological. An evaluation of agents that turn specifications into programs must test its instruments on planted errors, control for what models know without the specification, and replay exactly what it claims to check. Each of these changed a conclusion here, the last one through an external audit of our own scorer.

\textbf{Contributions.}
\begin{enumerate}[leftmargin=*,itemsep=1pt,topsep=2pt]
\item \textbf{Measurement validity.} LLM judges, a static analyzer and an LLM tester fail on planted mistakes, a description-free control exposes a conventions shortcut, and shared-interface scoring and adapted replay hide omissions. We give a fidelity metric with exact replay and explicit coverage, validated by known-answer tests (\S\ref{sec:method}, \S\ref{sec:rq1}, \S\ref{sec:rq5}).
\item \textbf{Information versus pipeline complexity.} Elicited answers raise fidelity across three model families, while the stages tested after generation add at most about one point and security rewriting removes required behavior; a factorial comparison separates information from architecture (\S\ref{sec:rq2}, \S\ref{sec:rq4}, \S\ref{sec:rq5}).
\item \textbf{Efficient elicitation.} How few questions suffice, how well agents choose and stop, how question value can be estimated without the original contract, and how synthetic description errors bias these answers (\S\ref{sec:rq3}).
\item \textbf{A resource.} The harness, 600 contract tasks with descriptions, client transcripts, reference models, every generated contract and every score, at \url{https://anonymous.4open.science/r/smart_contract_research_v2-ED3E/}.
\end{enumerate}

\begin{figure*}[t]
  \centering
  \includegraphics[width=\textwidth]{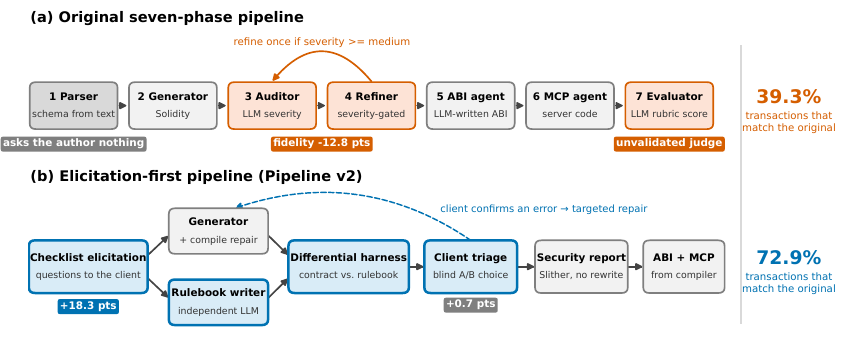}
  \caption{The two pipelines compared in this paper. (a) The original pipeline parses requirements without consulting their author, rewrites code in response to LLM security findings and grades itself with an LLM rubric. (b) The elicitation-first pipeline asks the client checklist questions before generation, checks the contract against an independently written reference model, shows each disagreement to the client as a blind choice and reports security findings without rewriting code. Badges give measured effects; percentages are transaction-level fidelity to the original contract on 500 held-out contracts (\S\ref{sec:rq5}).}
  \label{fig:pipeline}
\end{figure*}

%% ===================================================================
\section{Related Work}
%% ===================================================================

\textbf{Evaluating code generation.} Test-based benchmarks measure functional correctness against a trusted reference~\cite{chen2021evaluating, austin2021program}, and stronger test suites change model rankings~\cite{liu2023evalplus}. For Solidity, SolEval~\cite{peng2025soleval} and SolBench~\cite{chen2025solbench} test function completion, SolContractEval~\cite{ye2025solcontracteval} replays historical transactions and SolBehave~\cite{delvecchio2026solbehave} differentially fuzzes regenerated contracts against the originals. We also score against the original implementation, and our contribution is a study of the scoring rules and instruments themselves.

\textbf{LLMs as judges and auditors.} LLM judges are widely used and known to be biased~\cite{zheng2023judging}. For smart contracts, LLM auditors show high recall and low precision~\cite{chen2023chatgptvuln, sun2024gptscan}, and robust systems delegate the verdict to a non-LLM checker~\cite{xia2025symgpt, liu2025propertygpt}. Repair benchmarks report that security patches often break legitimate behavior~\cite{huang2026cyberchainbench, zhang2024acfix}.

\textbf{Oracles, clarification and user simulation.} LLM-written reference oracles~\cite{zhang2023algo}, consistency checks~\cite{sun2023clover, endres2023nl2postcond} and specification-guided differential testing~\cite{rao2024diffspec} verify code without ground-truth tests, building on differential and mutation testing~\cite{mckeeman1998differential, jia2011mutation}. ClarifyGPT~\cite{mu2023clarifygpt} and TiCoder~\cite{fakhoury2024ticoder} ask users clarifying questions, both with human evaluations. Simulated users are increasingly used to evaluate interactive systems, which raises questions about their validity~\cite{balog2025usersim}. Neither clarification nor executable oracles are new. We contribute a controlled study of how information, verification and repair affect behavioral fidelity of stateful programs, and of what such an evaluation can and cannot conclude.

%% ===================================================================
\section{Study Design}
\label{sec:method}
%% ===================================================================

\textbf{Corpus.} Contracts come from FSM-SCG~\cite{luo2025fsmscg}. Its implementations are not all reliable ground truth, since 35.1\% of the 21,976 contain placeholder markers and 27.8\% a comment-only function body. We keep non-stub contracts of at least 40 lines that compile at their declared Solidity version, deploy, and contain at least one contract-specific rule, meaning logic outside token and ownership plumbing. The development experiments use 100 such contracts and the scale replication 500 further contracts, none an exact duplicate after removing comments and whitespace; 12 share at least 90\% of their token 5-grams with a development contract, and excluding them moves no gap on the 500 contracts by more than 0.4 points. For each contract an LLM (Gemini 2.5 Flash) writes a short description of about 140 words, as a client might, and a complete requirements document of about 1,700 words that states every rule with exact values.

\textbf{Simulated client.} We use GPT-5.4 reading the real contract and answering each question in at most two sentences of plain English without mentioning code. Because the descriptions and the answers both derive from the original contract that fidelity is scored against, an elicitation gain measures how much of a hidden implementation a question recovers. It is an optimistic estimate of what a human author would provide under this simulator, not a measurement of it (\S\ref{sec:limits}).

\textbf{Executable reference models.} A writer model, called separately from the generator, receives the description, the contract's public interface and the deployment parameters, and returns a Python class with one method per public function. Methods take the caller first, raise \texttt{Revert} when the description says a call must fail and read the block time and attached ether from the harness. Models run in a separate process without imports and with restricted builtins; a reverted call restores the model's state and class-level state is reset between sequences, both covered by known-answer tests.\footnote{An earlier version of our runner did not restore state on revert. All reference-model results were re-graded after the fix; no strict catch rate moved by more than 0.6 points.} Reference models are fallible pipeline components. They are the object of the oracle-validity experiments, and inside the elicitation-first pipeline they flag candidate disagreements for the client to adjudicate; they never score generated contracts, which are always compared with the original Solidity contract.

\textbf{Transactions.} For each contract we generate 60 random sequences of 8 steps over its state-changing functions. Callers are drawn from three roles. Arguments mix small values, boundary values and values seeded from contract literals and getters with their neighbors; 15\% of steps advance the clock by an hour to a year; payable functions receive ether. Contracts run in an in-process EVM compiled at their declared version.

\textbf{Measure 1, oracle validity.} To test whether a reference model knows the required behavior, we plant one-rule mistakes into real contracts and keep an untouched twin~\cite{jia2011mutation}. A mistake is exposed if some transaction makes the twin and the mutant behave differently. Because outcomes are binary, disagreeing with the mutant somewhere is passed by a coin flip, so we use strict side-picking. Over exposing transactions we count how often the oracle sides with the twin and how often with the mutant, and the mistake is caught only if the former is larger. Two uninformed oracles, always succeed and always revert, expose class-specific shortcuts, and a description-free control writes the reference model from function names alone.

\textbf{Measure 2, implementation fidelity.} The original contract is the oracle for every fidelity number. We replay held-out sequences on it and on each generated contract, each continuing from its own state, so a step after a first divergence is still scored on its own. The required interface is every state-changing function of the original, the same for all systems. Replay is exact. Each step calls the original's signature with its argument types and attached ether, so a missing or changed signature fails the step and ether sent to a non-payable function reverts, and the chain clock is fixed so that every original scores 100\% against itself. A step is correct only if its outcome matches and, when it succeeds, every observed value changes as in the original; a change the generated contract does not expose is not credited. Observations cover up to 39 numeric, boolean and address-valued getters and the ether balance. Transaction-level fidelity is a contract's share of correct contract-specific steps, averaged over contracts; we also report the share of contracts that match the initial state and pass every sampled step, and the shares of required functions and value changes that can be observed. A system that never builds scores zero.

\textbf{Statistics and criteria.} Intervals are 95\% percentile bootstrap intervals over contracts with 2,000 resamples, paired for differences and conditional on one generation per condition, so they omit sampling variability. Each experiment states pass or fail criteria in its script; ten were committed together with their results and are marked as such (Appendix~\ref{app:prereg}). We test 36 criteria without correction for multiplicity, and we call an effect detectable when its interval excludes zero and meaningful when it also meets its criterion.

%% ===================================================================
\section{Results}
%% ===================================================================

\subsection{RQ1. Do common instruments measure intent?}
\label{sec:rq1}

\begin{table}[t]
\caption{Existing instruments on planted mistakes. Slither covers 200 guard removals, the LLM rows 40 mutants of 20 contracts and the reference-model rows 66 held-out contracts. A mistake is caught when the instrument flags the mutated function on the mutant but not on its twin, so an instrument that flags most contracts somewhere can still catch many. False alarms are the share of untouched contracts flagged, or for reference models the share of steps on untouched contracts where they disagree.}
\label{tab:instruments}
\centering\small
\setlength{\tabcolsep}{4pt}
\begin{tabular}{lcc}
\toprule
Instrument & Caught & False alarms \\
\midrule
Slither (static analysis) & 1.5\% [0.5, 4.3] & n/a \\
LLM judge reading the code & 65.0\% & 80\% \\
Black-box LLM tester & 17.5\% & 40\% \\
LLM judge with executable proof & 30.0\% & 40\% \\
\midrule
Reference model, with description & 94.3\% & 14.4\% \\
Reference model, function names only & 87.5\% & 18.7\% \\
\bottomrule
\end{tabular}
\end{table}

Table~\ref{tab:instruments} summarizes the audit. The static analyzer was built for vulnerability classes, not intent, and reported a new finding after only 3 of 200 guard removals. The LLM judge localizes many mistakes but flags 80\% of untouched contracts, and requiring an executable proof cut false flags but also true catches, leaving 30\%. The rubric judge of the original evaluation correlated with the aggregate of three human raters at $r{=}0.15$ on ten contracts, an interval of $[-0.53, 0.71]$, while the raters themselves agreed only moderately (ICC $0.33$). This is no evidence that it measured intent, the role it played.

Reference models catch 94.3\% of conventional mistakes such as removed modifiers and flipped comparisons. The more instructive number is the control. A reference model written from function names alone catches 87.5\%. Most of what a conventional benchmark rewards is knowledge of conventions that any capable model has, not understanding of the description. Each uninformed oracle also wins one mutation class, which is why we use strict side-picking, and the rest of the paper evaluates contract-specific rules.

\subsection{RQ2. Where is the information lost?}
\label{sec:rq2}

\begin{figure}[t]
  \centering
  \includegraphics[width=\columnwidth]{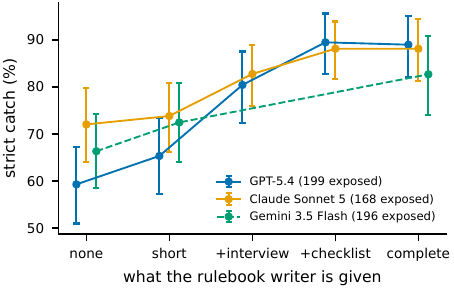}
  \caption{Strict catch of contract-specific mistakes by reference models, by what the writer is given, for three model families. Whiskers are 95\% intervals.}
  \label{fig:completeness}
\end{figure}

\textbf{Reference models.} On 100 fresh contracts with 279 contract-specific mistakes, a reference model written from the short description catches 65.9\% of exposed mistakes and one written from the complete requirements 87.5\%, a gap of 21.6 points $[15.1, 29.3]$, and 23.6 on fresh sequences (Figure~\ref{fig:completeness}). Counting mistakes no sequence exposes, the complete-description rate is 65.2\%. An extractor and a critic agent change nothing ($-1.0$), and the gap replicates on fresh sequences, generator ablations and two further model families. A checklist elicitor that asks ten fixed questions (Appendix~\ref{app:prompts}) recovers the full gap (89.4\% against 88.9\% for the complete document). A targeted interviewer that asks about whatever it judges unclear recovers only part of it (80.4\%), and a reviewer agent and a three-model ensemble applied to the complete document add nothing ($-2.5$, $0.0$). On planted mistakes, then, a reference model is reliable only when requirements are complete; \S\ref{sec:rq4} shows it is far weaker on real generation defects.

\textbf{Generated contracts.} Elicitation also helps the contracts a generator writes (GPT-5.4-mini, 92 contracts). Transaction-level fidelity is 55.6\% from the short description, 79.4\% with the interviewer's answers, 76.0\% with the checklist answers and 91.0\% from the complete document. Unlike for reference models, neither elicitor closes the gap, and the checklist does not beat the interviewer ($-3.4$ $[-8.3, 1.7]$). Checklist answers describe behavior only and leave argument types and stored values to the generator, so its contracts implement fewer required signatures (88.7\% against 94.8\% with the interviewer and 95.3\% from the complete document) and expose fewer of the original's values (78.6\% against 92.0\% and 98.6\%). The complete document also yields far more contracts that pass all sampled tests (59\% against 21\%). The gain does not depend on the generator. With Claude Sonnet~5 the checklist answers raise fidelity from 56.1\% to 75.1\% ($+19.0$ $[15.1, 23.0]$), and with Gemini~3.5~Flash from 51.8\% to 74.5\% ($+22.7$ $[17.7, 28.1]$).

\textbf{Is the gain an artifact of synthetic descriptions?} Our descriptions are written by an LLM from the real contract, so some contain false statements that elicitation then corrects. An auditor model (GPT-5.4) compared each short description with its contract. It found at least one contradiction in 76 of 100 descriptions, most often an access rule that does not exist (``an administrator can open trading'' when anyone can), and on average 14.9 omitted contract-specific rules. Elicitation therefore often corrects misinformation as well as filling omissions. On the 24 descriptions without any contradiction the checklist gain is still $+24.9$ $[14.1, 36.9]$ points, against $+18.8$ on the others, so the gain does not require contradictions, although with 24 contracts the two subgroups cannot be told apart ($+6.1$ $[-5.7, 18.3]$) and the audit is itself an LLM judgment. A simulated client generated from the answer can silently correct the inputs of the systems it evaluates~\cite{balog2025usersim}, and splitting results by audited input quality is a cheap check.

\textbf{An omission-only example.} The description of a liquidity-management contract, which the audit found free of contradictions, lists its states and operations but never says who may call them. Generated from it, the contract restricts pausing and resuming to an owner and makes withdrawals depend on an ether balance, matching 25\% of transactions. The single answer to ``anything unusual'' (all operations are open to any account, and no token accounting exists) raises fidelity to 100\%.

\subsection{RQ3. How many questions does the client need to answer?}
\label{sec:rq3}

\begin{figure*}[t]
  \centering
  \includegraphics[width=\textwidth]{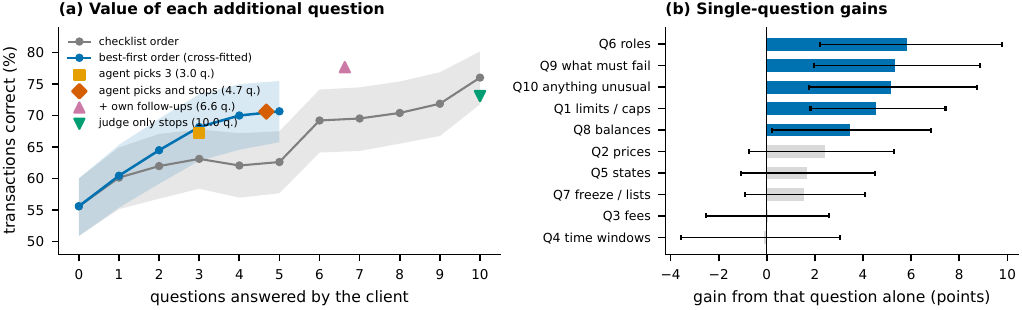}
  \caption{Question budget (92 contracts, transaction-level fidelity). (a) Asking the checklist in its fixed order gains little until question 6. Ordering questions by their measured value, learned on the other half of the contracts, reaches 68\% with three questions and 71\% with five. Points show adaptive agents at their mean number of questions. (b) Gain from each question asked alone; a few questions carry most of the value. Bands and whiskers are 95\% intervals.}
  \label{fig:budget}
\end{figure*}

Reusing the client's answers, we generated contracts from every prefix of the checklist, from each question alone, from questions ordered by their single-question value and from four adaptive agents (Figure~\ref{fig:budget}). Ordering is cross-fitted, learned on one half of the contracts by hash and applied to the other. Because answers come from one session, a subset only approximates a shorter conversation.

\textbf{A few questions carry most of the value.} Alone, the questions about roles ($+5.8$ $[2.2, 9.8]$), failure conditions ($+5.3$), anything unusual ($+5.2$) and caps ($+4.5$) gain most; fees and time windows add nothing on average and prices little. Ordered best first, three questions reach 68.2\% and recover about 62\% of the ten-question gain and five reach 70.7\% and recover about 74\%, while the first three checklist questions recover 37\%. The valuable questions are the ones about permissions and failures, which is exactly where the description audit found most contradictions. The roles question gains 7.3 points $[1.5, 13.1]$ on the 52 contracts without explicit caller checks and 4.0 on the 40 with them, because a generator left to its conventions adds owner-only restrictions that a permissive contract does not have. This pattern partly reflects the synthetic descriptions. On the 24 descriptions without contradictions, caps, balances, freezes and anything unusual each gain 8 to 10 points alone, roles $+6.9$ $[-0.1, 15.0]$, and the best-first order learned on all contracts does not beat the checklist order (66.4\% against 72.7\%), so question order should be learned on inputs like those it will serve.

\textbf{Agents do not stop well.} An agent that picks three questions per contract (67.2\%) is 4.1 points above the first three checklist questions $[-1.0, 8.9]$, short of its criterion, and most often asks about roles first. An agent that also decides when to stop asks 4.7 questions and reaches 70.6\%, 5.4 points below asking all ten $[-9.6, -1.7]$, and a judge that only decides sufficiency never stopped early. An agent allowed to write its own follow-ups asks 6.6 questions and reaches 77.7\%, 7.1 points $[2.8, 11.7]$ above the stopping agent, with the highest interface coverage of the question agents (90.2\%). LLM judgments help choose and phrase questions but are not calibrated for sufficiency.

\textbf{Question value can be estimated without the original contract.} For each question we measured how often the contract written with its answer behaves differently from the one written without it. This separation score needs the client's answer but not the original contract, and it ranks the ten questions much as their measured gain does (Spearman $\rho{=}0.92$). It can order the checklist offline from earlier clients' answers. Ordering by mean separation on one half of the contracts, with test sequences built from the generated contracts so that the original never enters, selects in both halves the same three questions that the label-based order selects (anything unusual, roles and failure conditions). Asked first on the other half, they reach 66.7\%, not reliably above the first three checklist questions ($+3.6$ $[-1.6, 8.5]$) and $-1.5$ $[-5.7, 2.3]$ from the label-based order, which differs only in answer order; both criteria fixed at three questions are missed. At five questions the label-free order beats the checklist by 7.8 points $[3.3, 12.2]$ and matches the label-based order ($-0.3$ $[-2.6, 2.3]$).

\subsection{RQ4. What do the pipeline's stages do?}
\label{sec:rq4}

\begin{figure*}[t]
  \centering
  \includegraphics[width=\textwidth]{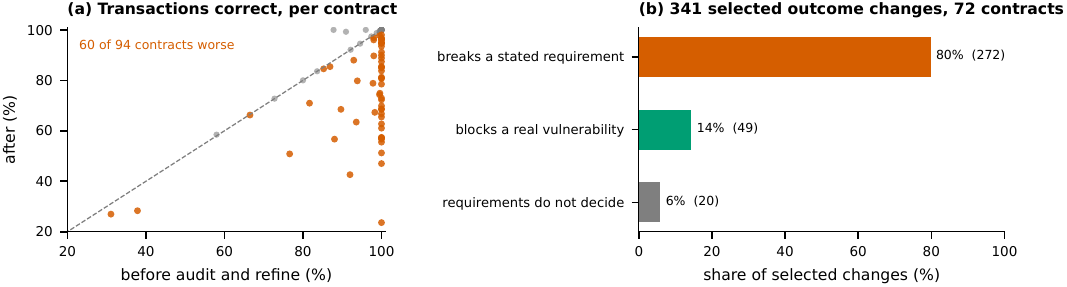}
  \caption{What the security audit-and-refine loop does. (a) Transaction-level fidelity per contract before and after the loop, with complete requirements as input; points below the diagonal lost fidelity. (b) LLM classification of a capped selection of 341 outcome changes in the 72 rewritten contracts against the complete requirements.}
  \label{fig:security}
\end{figure*}

\textbf{Security-driven rewriting.} With complete requirements as input, generation alone reaches 95.3\% fidelity on 94 contracts. The original pipeline's audit then rates most contracts high or critical, its refiner rewrites 86, and fidelity falls to 82.5\%, a loss of 12.8 points $[9.5, 16.3]$ with 60 contracts worse and the share passing all sampled tests falling from 62\% to 23\%. Fidelity and safety are separate outcomes, and a change that violates an unsafe specification can still be justified, because the original contracts can be permissive. We therefore had GPT-5.4 classify a capped selection of behavior changes against the complete requirements, 341 deduplicated success-or-revert changes in the 72 rewritten contracts, at most six per contract, drawn from 4,886 changed steps. Of the selected changes, 80\% break a stated requirement, 14\% block a genuine vulnerability that the requirements do not ask for and 6\% are not decided by the requirements. These labels are not human-verified. On the 51 contracts with no justified fix among the selected changes, fidelity falls by 17.4 points. The loop mostly removes behavior the client asked for rather than trading it for safety (Figure~\ref{fig:security}); 44 of the 49 justified fixes add a restriction that could have been put to the client as a question.

\textbf{Reference-model verification.} Repairing contracts where they disagree with a reference model has no detectable effect ($-0.2$ $[-0.8, 0.2]$) when requirements are complete. The instrument's validity does not transfer from planted to real mistakes. The complete-description model catches 94.3\% of planted conventional and 87.5\% of exposed contract-specific mistakes, but on real generation defects, the 3.2\% of steps where a generated contract differs from the original, it sides with the original only 31.6\% of the time on a common set of 1,217 outcome disagreements, against 80.4\% for always succeed. Most residual defects are conservative checks the description never mentions, which a model written from it repeats. Only 34.8\% of its flags mark real defects, an elevenfold enrichment over the 3.2\% base rate but wrong more often than right. Description-derived checkers verify what a description states and cannot adjudicate its silences, so their flags go to the client as questions.

\begin{figure}[t]
  \centering
  \includegraphics[width=\columnwidth]{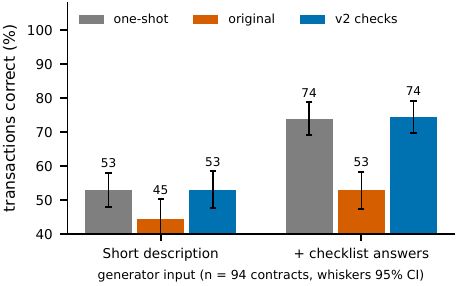}
  \caption{Information versus architecture. Each architecture receives either the short description or the same description with the checklist answers the elicitation-first pipeline obtained.}
  \label{fig:factorial}
\end{figure}

\textbf{Information versus architecture.} A better pipeline may simply receive more information, so we ran each architecture on both inputs (Figure~\ref{fig:factorial}, 94 contracts). The answers lift a one-shot generator by 21.0 points but the original pipeline by only 8.4 $[3.3, 13.2]$, so given identical information the original pipeline ends 21.0 points below a one-shot generator and 21.5 below the elicitation-first pipeline. Verification and client triage without the answers show no detectable effect ($+0.1$ $[-3.2, 3.2]$) and with them a detectable but small 0.5 points $[0.1, 0.9]$. The interaction is large ($+13.0$ $[7.4, 19.3]$). Information is necessary, and the original architecture wastes most of it, largely through the rewriting described above.

\subsection{RQ5. Does it hold at scale, and does the metric matter?}
\label{sec:rq5}

\begin{table}[t]
\caption{End to end on 500 held-out contracts (499 with scorable steps), all from the same short description and code generator. Never-built outputs score zero. Columns give transaction-level fidelity, contracts passing all sampled tests, contracts that build, required functions implemented and original value changes observable.}
\label{tab:scale}
\centering\small
\setlength{\tabcolsep}{3.5pt}
\begin{tabular}{lccccc}
\toprule
& Txn & Pass & Build & Func. & Obs. \\
\midrule
One-shot generation & 53.9 & 2.6 & 95.2 & 75.5 & 75.6 \\
Original pipeline & 39.3 & 1.2 & 79.2 & 58.7 & 67.1 \\
Elicitation-first, before triage & 72.2 & 17.2 & 97.0 & 84.7 & 77.7 \\
Elicitation-first, full & \textbf{72.9} & \textbf{19.6} & 97.0 & 84.7 & 77.7 \\
\bottomrule
\end{tabular}
\end{table}

\textbf{Scale.} On 500 held-out contracts (Table~\ref{tab:scale}), the elicitation-first pipeline exceeds the original pipeline by 33.5 points $[30.7, 36.5]$ and one-shot generation by 18.9 $[16.6, 21.3]$, and the original pipeline falls 14.6 points below one-shot generation. The comparison with one-shot generation is the informative one. The original pipeline parses the description into a schema, generates, runs an LLM auditor whose findings rated medium or higher trigger one refinement round, and adds ABI, MCP and evaluator agents (Figure~\ref{fig:pipeline}a). On the 381 contracts where every system builds it still trails one-shot generation by 7.9 points, so about half of its deficit comes from outputs that never build and half from behavior. Triage adds a detectable 0.7 points $[0.3, 1.2]$, below its pre-set threshold and provisional, because the pipeline's verification stage ran before the runner fix; elicitation accounts for about 96\% of the gain over one-shot generation. Even the best pipeline delivers contracts that pass all sampled tests in only 20\% of cases and implements 85\% of the required functions.

\begin{figure*}[t]
  \centering
  \includegraphics[width=\textwidth]{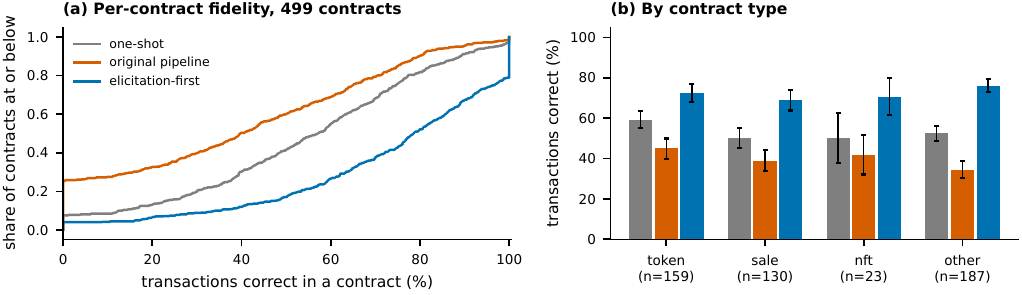}
  \caption{Beyond the averages. (a) Cumulative distribution of per-contract transaction-level fidelity on 499 contracts; a curve further to the right is better, and the vertical jump at 100\% is the share of contracts with every step correct. (b) Mean fidelity by contract type with 95\% intervals; colors as in (a).}
  \label{fig:distribution}
\end{figure*}

\textbf{Distributions.} The original pipeline scores zero on 25.5\% of contracts, mostly because nothing it delivers builds (Figure~\ref{fig:distribution}). One-shot generation has a median of 57.1\%. The elicitation-first pipeline has a median of 78.6\%, 32.9\% of contracts at or above 90\%, and beats the original on 424 contracts and loses on 46. Its gain is largest on contracts that are not tokens, sales or NFTs, where the original mostly fails by omission (55\% against 87\% of required functions).

\textbf{What still goes wrong.} Of the steps the elicitation-first pipeline gets wrong, 45\% call a function the contract lacks or declares with different argument types, 29\% have the wrong outcome, 6\% the wrong change in a balance, counter or state, and 20\% change a value the contract does not expose; 202 of 499 contracts lack at least one required signature. Descriptions name operations but not argument types, so much of this is interface guessing that a typed interface in the request would remove.

\begin{figure*}[t]
  \centering
  \includegraphics[width=\textwidth]{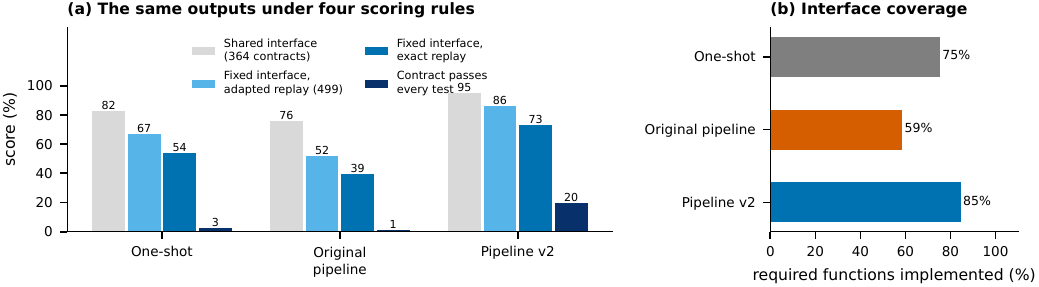}
  \caption{The same 500 sets of outputs scored four ways. (a) The shared-interface metric drops any function some system lacks and excludes contracts that do not build; adapted replay calls each contract through its own signatures; exact replay calls the original's. (b) Share of the original contract's contract-specific functions each system implements.}
  \label{fig:metric}
\end{figure*}

\textbf{Metric sensitivity.} The earlier evaluation of this pipeline~\cite{priorbenchmark2026}, like many comparisons, scored only functions all systems implement on contracts where all build. Figure~\ref{fig:metric} scores the same outputs four ways. The shared-interface metric reports 95\% for the elicitation-first pipeline and 76\% for the original, a gap of 19 points on 364 contracts. With a fixed required interface the gap is 35 points on 499 contracts, and 34 under exact replay. Restricting to a shared interface removes 28\% of the original's contract-specific functions in a typical comparison and with them the failures of the system that omitted them, and excluding contracts where any system fails to build removes 27\% of the sample, mostly failures of the original pipeline.

\textbf{Auditing our own scorer.} An external audit of the previous version of our scorer found four ways a wrong contract could score 100\%. Calls were adapted to the generated contract's own signatures, ether was dropped for non-payable functions, a value change the generated contract did not expose was skipped, and only changes, not initial values, were compared. Under that version 43.9\% of the elicitation-first pipeline's contracts passed every test, but 162 of those 219 had an unobserved value or a missing or mismatched function. The scorer used throughout this paper replays exactly and passes known-answer tests built from all four counterexamples. It lowers fidelity by 12 to 14 points for every system and the pipeline's pass rate to 19.6\%, yet moves no gap between systems by more than 1.2 points. Every rule we tried preserves the ranking, but absolute fidelity, gap sizes and pass rates depend on the rule, so all should be reported with interface and observation coverage.

\subsection{Robustness of the main conclusions}
\label{sec:robust}

\begin{table}[t]
\caption{Each main conclusion and the independent checks behind it.}
\label{tab:robust}
\centering\small
\setlength{\tabcolsep}{3pt}
\begin{tabular}{p{2.6cm}p{5.3cm}}
\toprule
Conclusion & Checks that support it \\
\midrule
Information is lost before code & fresh transaction sequences; five transaction-generator variants; three rulebook writer families; generated contracts as well as rulebooks; descriptions without factual errors; three generator families \\
\addlinespace
Stages after writing add at most about one point & no detectable gain from extractor, critic, reviewer, ensemble, voting, proof-gated judging or rulebook repair; triage detectable but small and provisional \\
\addlinespace
Security rewriting breaks requirements & 94 contracts; LLM classification of selected changes; holds on contracts without any selected justified fix \\
\addlinespace
Elicitation-first beats the original & 96 and 499 contracts; four contract types; equal-information factorial; shared, adapted and exact scoring; failures counted or excluded; near-duplicates excluded \\
\addlinespace
Few questions recover much of the gain & cross-fitted ordering; four adaptive agents; a label-free separation score; which questions matter depends on description errors \\
\addlinespace
Instrument fixes do not drive results & runner fix moved no catch rate by more than 0.6 points; exact replay preserved every ranking \\
\bottomrule
\end{tabular}
\end{table}

Table~\ref{tab:robust} lists, for each conclusion, the checks that would have contradicted it had it been an artifact, varying the instrument (sequences, generator ablations, scoring rules, runner fixes) or the population (contracts, types, model families). Conclusions that did not survive are reported as failed criteria.

%% ===================================================================
\section{Discussion}
%% ===================================================================

\textbf{Lessons for evaluating generative pipelines.} Several evaluation choices changed our conclusions, and reporting failures mattered as much. Eighteen of our criteria failed, and consistently so. No agent applied after the requirements were fixed produced a meaningful gain, targeted questioning never beat the generic checklist by its margin, and adaptive stopping missed its margin. We would ask the following of any evaluation of systems that turn natural-language requirements into executable artifacts.
\begin{enumerate}[leftmargin=*,itemsep=1pt,topsep=2pt]
\item Validate every instrument on planted errors with untouched twins, and report its false alarms next to its detections.
\item Include a control that withholds the specification, and report how much of a score survives without it.
\item Score a fixed required interface by exact replay, count outputs that do not build as failures, report coverage and pass rates, and test the scorer on known wrong answers.
\item When users or inputs are simulated from the answer, audit the inputs, report results separately for inputs the simulator could have repaired, and treat the gains as optimistic.
\item Fix pass and fail criteria before running, and report the failures with the same prominence as the successes.
\end{enumerate}

\textbf{What the results say about pipeline design.} In the configurations we tested, information gathered before generation mattered far more than anything after it. A few broad questions recover much of it, and verification is best used as an alarm that enriches real defects but is usually wrong, so its disagreements go back to the client. Automated security repair should turn its findings into questions rather than rewrite. Architecture matters mostly by not destroying information.

\textbf{Implications for decentralized and Web applications.} Smart contracts are the logic layer of decentralized Web applications, so these findings bear directly on how contract-writing agents for them should be evaluated. We hypothesize, without having tested it, that Web agents that act on a user's request share the same problems, a request that underdetermines behavior, judges that stand in for users and simulated users that can leak the answer, and that replayable sandboxes, description-free controls and scored clarification would transfer to them.

%% ===================================================================
\section{Limitations}
\label{sec:limits}
%% ===================================================================

\textbf{Construction and simulated client.} Descriptions and answers are generated from the contract that fidelity is scored against, so we measure implementation fidelity under incomplete information, not author intent. No human answered our questions; real authors may be unsure or inconsistent, or may volunteer more, so elicitation gains are optimistic estimates under this simulator. We had no independently authored specification or expert-written tasks, and no controlled omission-only condition. \textbf{Originals as intent.} We treat the original contract as the intended behavior even where it lacks access control. The security classification addresses this, but its labels come from an LLM, cover a capped selection of outcome changes and were not verified by people. \textbf{Coverage.} Observations exclude events and ether received by external accounts, and random sequences do not expose every rule. \textbf{Baselines and scope.} The pipeline baseline is our earlier system and one-shot generation, not ClarifyGPT or TiCoder. All contracts come from one corpus and one language, generator families share a compile-repair step, and criteria are not corrected for multiplicity. The pipeline's verification stage ran before the reference-model runner was fixed and was not rerun, so its sub-point benefit is provisional; intervals omit generation-to-generation variability.

%% ===================================================================
\section{Future Work}
%% ===================================================================

\textbf{Human authors.} The most important next step is a study with people who commission contracts and write their requirements before any implementation exists, which would measure intent rather than implementation fidelity and the effort that answering questions costs a person. \textbf{Autonomous agents that transact.} LLM agents increasingly hold wallets and call contracts for their users, paying, trading or settling escrow, and agents may negotiate and transact with one another. The same replayable sandbox could evaluate such agents end to end, scoring whether each transaction an agent sends matches its user's request, whether the agent that writes a contract and the agents that use it read it the same way, and when an agent should ask its user before moving funds. \textbf{Security as questions.} A security agent that phrases each finding as a question to the client would keep the justified fixes without the regressions. \textbf{Broader settings.} A second corpus, comparisons with published clarification systems, repeated generations and other languages would test how far the findings generalize.

%% ===================================================================
\section{Conclusion}
%% ===================================================================

Evaluating agentic code synthesis requires instruments that are themselves tested, controls for what models know without the requirements, a scorer that replays exactly and does not hide omissions, and checks on the simulated users that stand in for authors. Each of these changed a conclusion in the earlier evaluation of this pipeline or in our own. On measurement, LLM judges, static analysis and LLM testers failed on planted mistakes, a description-free control showed that conventional benchmarks largely reward conventions, and shared-interface scoring and adapted replay hid omissions until exact replay and known-answer tests exposed them. On pipeline design, information supplied before generation mattered far more than anything after it. Elicited answers raised fidelity by 19 to 23 points across three generator families, while review, repair and triage added at most about one point and security rewriting removed required behavior. On elicitation, three to five well-chosen questions recovered 60 to 75\% of what a simulated client could supply, and question value could be ranked without the original contract, although which questions matter depended on errors in the synthetic descriptions. A pipeline rebuilt on these findings reaches 72.9\% implementation fidelity on held-out contracts against 53.9\% for one-shot generation and 39.3\% for the original, yet only 20\% of its contracts pass all sampled tests, a gap that studies with human authors should address next. All code, tasks, client transcripts and scores are released (Appendix~\ref{app:resource}).

\bibliographystyle{ACM-Reference-Format}
\bibliography{references}

\appendix
%% Appendix for main.tex (included with \input{appendix}).

\section{System Architecture and Codebase}
\label{app:arch}

\begin{figure*}[t]
  \centering
  \includegraphics[width=\textwidth]{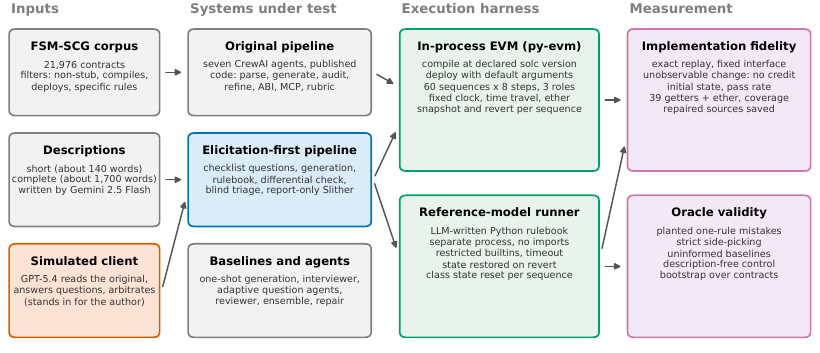}
  \caption{Architecture of the study. Inputs (left) feed the systems under test; every system's output runs in the same in-process EVM, and reference models run in a separate sandboxed process. Two measurement families consume the traces, fidelity of generated contracts and validity of oracles on planted mistakes.}
  \label{fig:codebase}
\end{figure*}

Figure~\ref{fig:codebase} shows how the parts fit together and Table~\ref{tab:modules} maps them to the released code. Every model request and output is cached on disk by contract and condition, so each stage is resumable and every reported number can be recomputed without model calls.

\begin{table}[h]
\caption{Main modules of the released code.}
\label{tab:modules}
\centering\small
\setlength{\tabcolsep}{3pt}
\begin{tabular}{@{}>{\raggedright\arraybackslash}p{3.0cm}>{\raggedright\arraybackslash}p{5.0cm}@{}}
\toprule
Module & Role \\
\midrule
\texttt{harness.py} & deploy at the declared compiler version, run steps with callers, ether and time travel \\
\texttt{runner\_state.py}, \texttt{model\_runner3.py} & sandboxed reference-model execution with revert rollback and per-sequence reset \\
\texttt{idea8\_specific.py} & transaction generator, seed pool, contract-specific mutation operators \\
\texttt{rbeval.py} & strict side-picking, right-side rate, bootstrap over contracts \\
\texttt{fidelity.py}, \texttt{rescore.py} & fidelity with exact replay, saved compile repairs, real-defect analysis \\
\texttt{exp*.py}, \texttt{tests/} & one script per experiment with its criteria; known-answer tests \\
\texttt{pipeline/v2}, \texttt{demo/} & elicitation-first pipeline with command-line and web interfaces \\
\texttt{out/} & every cached model output, saved repair and per-contract score \\
\bottomrule
\end{tabular}
\end{table}

\section{Pipeline Algorithm}
\label{app:algo}

The elicitation-first pipeline is short. Model calls are marked; everything else is deterministic.

\noindent\begin{minipage}{\columnwidth}
\begin{lstlisting}
def pipeline(desc, name, client):
  ans = client.answer(CHECKLIST)     # client
  spec = desc + transcript(ans)
  code = generate(spec, name)        # model
  code = compile_repair(code)        # model
  ref = write_model(spec, abi(code)) # model
  for rnd in (1, 2):
    cases = diff(code, ref, rnd)     # EVM vs ref
    if not cases: break
    picks = client.choose(cases)     # blind A/B
    bad = [c for c in cases
           if picks[c] == "ref"]
    if bad:
      code = repair(spec, code, bad) # model
    ref = write_model(spec + picks,
                      abi(code))     # model
  report = slither(code)    # reported only
  abi_, mcp = from_compiler(code)
  return code, ref, report, abi_, mcp
\end{lstlisting}
\end{minipage}

\begin{table*}[t]
\caption{Predefined criteria and outcomes, in the order the experiments were run. Criteria marked $\S$ were committed together with their results and are retrospective; the others were committed before any output existed. Reference-model criteria ($\dagger$) were re-graded with the corrected runner and none changed outcome. Criteria for generated contracts were set on an earlier metric and are applied unchanged to the exact-replay metric.}
\label{tab:prereg}
\centering\footnotesize
\setlength{\tabcolsep}{4pt}
\begin{tabular}{llll}
\toprule
Experiment & Criterion & Outcome & Value \\
\midrule
Slither on guard removal$^\S$ & detection below 25\% & met & 1.5\% \\
Conventional over-permission$^\S$ & violation rate at least 5\% & not met & 0.2\% \\
Tester versus judge$^\S$ & tester at least judge plus 15 & not met & 17.5 vs 65.0 \\
Proof-gated judge$^\S$ & recall at least 50\% & not met & 30\% \\
Implementation voting$^\S$ & majority above mean single & not met & 80.5 vs 80.9 \\
Disagreement-guided clarification$^\S$ & $+5$ over none, $+3$ over random & not met & $+1.7$ \\
Reference model pilot$^{\dagger\S}$ & right side at least 75\%, beats uninformed & met & 21 of 23 \\
Reference models at scale$^{\dagger\S}$ & description adds 10 over function names & not met & $+6.8$, $0.0$ \\
Contract-specific, short$^{\dagger\S}$ & description adds 10 over function names & not met & $+6.0$, $+9.0$ \\
Extractor and critic$^{\dagger\S}$ & $+10$ over complete document & not met & $-1.0$ \\
Robustness, fresh sequences$^\dagger$ & complete minus short at least $+15$ & met & $+23.6$ \\
Gemini and Claude replication$^\dagger$ & complete minus short at least $+10$ & met & $+10.2$, $+14.3$ \\
Interviewer$^\dagger$ & recovers at least half the gap & met & $+15.1$ of $23.6$ \\
Targeted versus generic (models and contracts) & targeted at least $+5$ and $+3$ & not met & $-9.0$; $+3.4$, spans zero \\
Reviewer and ensemble$^\dagger$ & at least $+3$ & not met & $-2.5$, $0.0$ \\
Generator ablation$^\dagger$ & gap at least $+10$ under every variant & met & $+20.3$ to $+21.7$ \\
Rulebook repair versus generation & at least $+3$ & not met & $-0.2$ \\
Rulebook repair versus security loop & at least $+3$ & met & $+12.6$ \\
Interviewer, generated contracts & at least $+3$ and half the gap & met & $+23.8$ of $35.4$ \\
Real generation defects & right side at least 75\% & not met & 31.6\% \\
Pipeline versus original (96 and 499 contracts) & at least $+5$ & met & $+29.2$, $+33.5$ \\
Pipeline versus one-shot (96 and 499 contracts) & at least $+5$ & met & $+20.9$, $+18.9$ \\
Original below one-shot (499) & interval below zero & met & $-14.6$ \\
Triage over elicitation (96 and 499) & at least $+1$, interval above zero & not met & $+0.5$, $+0.7$ \\
Original pipeline gains from answers (F1) & at least $+5$ & met & $+8.4$ \\
Architecture at equal information (F2) & at least $+3$ & met & $+21.5$ \\
Verification without answers (F3) & at least $+3$ & not met & $+0.1$ \\
Gain on error-free descriptions (D2) & at least $+5$ & met & $+24.9$ \\
Justified security fixes (S2) & below 25\% of selected changes & below & 14\% \\
Question budget, checklist order (H-B1) & 90\% of gain within 5 questions & not met & needs 10 \\
Best-first order (H-B2) & at least $+2$ over checklist at 3 & met & $+5.1$ \\
Adaptive stopping (H-B3) & at most 5 questions, within 2 points & not met & 4.7 q., $-5.4$ \\
Agent selection (H-B4) & at least $+2$ over checklist at 3 & not met & $+4.1$, spans zero \\
Free follow-ups (H-B5) & at least $+1$ over adaptive & met & $+7.1$ \\
Generator families (G1) & checklist gain at least $+5$ per family & met & $+19.0$, $+22.7$ \\
Label-free order (L1, L2) & $+2$ over checklist, within 2 of label-based & not met & $+3.6$, $-1.5$ \\
\bottomrule
\end{tabular}
\end{table*}

\section{Measurement Details}
\label{app:measure}

\textbf{Oracle validity.} For a mutant $m$ of contract $c$ with exposing transactions $T_m$, an oracle $O$ collects
$s^+_O(m) = |\{t \in T_m \mid O(t) = c(t)\}|$ and $s^-_O(m) = |\{t \in T_m \mid O(t) = m(t)\}|$.
The mutant is strictly caught when $s^+_O(m) > s^-_O(m)$, and the right-side rate is $\sum_m s^+_O(m) / \sum_m (s^+_O(m) + s^-_O(m))$. Outcome and value-change disagreements both count, except in the conventional experiment, which uses outcomes only. Catch rates are reported over exposed mutants; over all mutants, the complete-description rate is 65.2\% (182 of 279) rather than 87.5\%.

\textbf{Fidelity.} Let $S$ be the set of steps on contract-specific functions at which the original contract $c$ executes with a valid outcome. Each step calls the original's signature with its argument types and ether value; a generated contract $g$ without that signature cannot execute it, and ether sent to a non-payable function reverts. A step $s \in S$ is correct for $g$ when $g(s) = c(s)$ and, if both succeed, $\Delta_g(s, o) = \Delta_c(s, o)$ for every observation $o$ whose change is defined for $c$; a nonzero change in $c$ that cannot be observed on $g$ makes the step unverified, and unverified steps are not credited. Transaction-level fidelity is the share of correct steps in $S$, computed per contract and averaged over contracts, so every contract weighs equally; differences between systems are bootstrapped over the same contracts. Each contract replays every sequence from its own state, so after a first divergence its later steps are still compared one by one, and a step can be correct after an earlier wrong one; the pass criterion, every step correct, is the strict sequence-level summary. A contract passes when every step in $S$ is correct and no value readable on both contracts differs after deployment, the ether balance excepted; deltas alone could hide a wrong initial state, which this check catches when both constructors take the same parameters. Interface coverage is the share of the original's contract-specific functions present in $g$ with the same name and argument types, and observation coverage the share of the original's value changes observable on $g$; the observation set is fixed by the original contract. The chain starts at a fixed timestamp and each block adds one second, so replay is deterministic, and an original scored against itself passes. Known-answer tests cover a missing getter, a wrong initial value, a changed signature and ether sent to a non-payable function, each of which the previous scorer passed.

\begin{table}[h]
\caption{Transaction generator and statistics.}
\label{tab:params}
\centering\small
\setlength{\tabcolsep}{3pt}
\begin{tabular}{@{}>{\raggedright\arraybackslash}p{2.9cm}>{\raggedright\arraybackslash}p{5.1cm}@{}}
\toprule
Setting & Value \\
\midrule
Sequences per contract, steps & 60, 8 \\
Callers & owner (deployer), two other accounts \\
Numeric arguments & 0, 1, 100, 999, 1000, 1001, seeded values \\
Seed pool & contract literals and getter values $\pm 1$, at most 16 \\
Time steps & 15\% of steps, 1 hour to 366 days \\
Clock & fixed start, one second per block \\
Ether sent to payable calls & 0, $10^{15}$ to $10^{18}$ wei, seeded values \\
Observations & up to 39 getters and the ether balance \\
Bootstrap & 2,000 resamples over contracts \\
\bottomrule
\end{tabular}
\end{table}

\section{Additional Results}
\label{app:results}

\begin{figure}[h]
  \centering
  \includegraphics[width=\columnwidth]{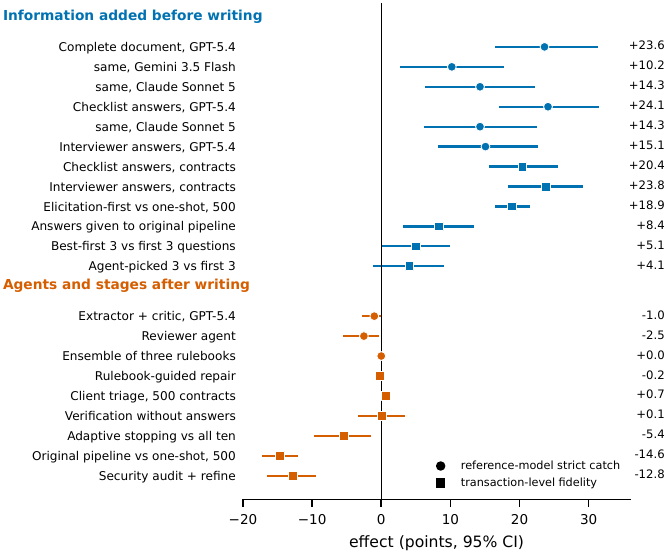}
  \caption{Every intervention in the study, by where it acts. Interventions that add information before writing (blue) help in every model family and on both measures; agents and stages acting after writing (red) add at most about one point or cause harm. The first row uses the fresh-sequence evaluation.}
  \label{fig:effects}
\end{figure}

\textbf{Complete picture of effects.} Figure~\ref{fig:effects} collects every intervention. No intervention that adds information before writing has an interval below zero, and the largest positive effect of any stage acting after writing is client triage at under one point.

\textbf{By mutation class and false alarms.} With the GPT-5.4 writer, checklist answers and the complete document lift the catch rate in every class of contract-specific mistake, and agents applied to the complete document leave every class unchanged or lower. The same information lowers disagreement with untouched contracts from 21.9\% to 4.2\% of transactions, so a reference model written from a short description is also noisier.

\begin{table}[h]
\caption{Behavior of the adaptive question agents (92 contracts).}
\label{tab:agents}
\centering\small
\setlength{\tabcolsep}{3pt}
\begin{tabular}{lccl}
\toprule
Agent & Questions & Fidelity & Most frequent first question \\
\midrule
Picks three & 3.0 & 67.2\% & roles (56 of 93) \\
Picks and stops & 4.7 & 70.6\% & roles (46 of 93) \\
Picks, stops, own questions & 6.6 & 77.7\% & roles (44 of 93) \\
Judge only stops & 10.0 & 73.1\% & never stopped early \\
\midrule
All ten questions & 10 & 76.0\% & \\
\bottomrule
\end{tabular}
\end{table}

\textbf{Question agents.} Table~\ref{tab:agents} summarizes the agents. All of them open with the roles question most often, the single most valuable one. The agent allowed to write its own questions wrote 260 follow-ups across 93 contracts and is the only agent that matches asking all ten.

\textbf{Client triage and effort.} In the 500-contract run the client saw 2,207 disagreements and confirmed the reference model in 53\%, the contract in 32\% and neither in 14\%, so neither side can be trusted by default. Besides answering ten questions, the client judged 8.5 situations on average in the 53\% of contracts where disagreements reached it, which for a person would be the dominant cost. The pipeline makes 3.8 model calls per contract besides the client's answers, with a median wall time of 107 seconds (mean 989, inflated by rate limits).

\section{Predefined Criteria}
\label{app:prereg}

Table~\ref{tab:prereg} lists all 36 criteria; the ten marked $\S$ entered the repository with their results, the others before any output. Eighteen were not met. The main text reports effects with intervals, detectable when the interval excludes zero and meaningful when the criterion is also met; no correction for multiplicity is applied.

\section{Prompts}
\label{app:prompts}

\textbf{Description writer.} Writes the 120 to 200 word description a product owner would write, naming operations but without code; the complete document states every rule with exact numbers and boundaries.

\textbf{Generator.} Implements every described operation as one self-contained Solidity 0.8.20 contract with the given name, without imports.

\textbf{Reference-model writer.} Given the requirements, public interface and deployment arguments, writes \texttt{class Model} with one method per function, sender first, raising \texttt{Revert} when a call must fail and choosing the most conventional behavior where the requirements are silent.

\textbf{Checklist questions.} (1) limits or caps on amounts, with exact values and inclusiveness; (2) prices, rates or required payments; (3) fees and how they are calculated; (4) time windows, deadlines or waiting periods; (5) phases or states and the operations each allows; (6) operations restricted to accounts or roles, and who holds them; (7) freezing, pausing, blacklisting or whitelisting; (8) how balances or supply change per operation; (9) which operations should fail and when; (10) anything unusual a developer might get wrong.

\textbf{Simulated client.} Reads the original contract and answers each question in at most two plain-English sentences with exact numbers, describing behavior only and nothing beyond the question. For triage it sees two options in random order and answers 1, 2 or 0 for neither.

\section{Case Study and Transcript}
\label{app:case}

The short description of an NFT key contract says an administrator changes the sale status and adds addresses to the gold list, but the real contract has no privileged role. One-shot generation and the original pipeline implement the conventional reading; the answer to the roles question corrects it.

\begin{itemize}[leftmargin=*,itemsep=1pt]
\item \emph{Which operations are restricted to specific accounts or roles?} ``No operation is restricted to a privileged owner, admin, or special role. Any account may add any address to the gold list, change the sale status [...] and mint only if its own address is on the silver or gold list and the sale is live.''
\end{itemize}

The contract generated after all ten questions agrees with the original on every outcome and every value it exposes, but 22\% of steps change a value it does not expose, so exact scoring credits it with 77.6\%, against 41.8\% for one-shot generation and 22.4\% for the original pipeline. Unlike the example in \S\ref{sec:rq2}, this is a description error.

\section{Resource Documentation}
\label{app:resource}

\textbf{Contents.} For each of 600 contracts (100 development, 500 replication) the release contains the original source, the short description, every generated or repaired contract and per-contract scores; the development contracts also have complete requirements, client answers, interviewer transcripts and reference models. Original-pipeline records keep the final contract and a log tail, and its code is not redistributed.

\textbf{Models.} Descriptions were written by Gemini 2.5 Flash. Reference models were written by GPT-5.4, Gemini 3.1 Pro (preview), Gemini 3.5 Flash, Claude Sonnet 5 and, in the pilot, Claude Sonnet 4.5. Contracts were generated by GPT-5.4-mini, and in the replication by Claude Sonnet 5 and Gemini 3.5 Flash (9 of 372 requests returned nothing usable and count as never built), with the same GPT-5.4-mini compile repair for every family. The client, auditor, security classifier and question agents use GPT-5.4.

\textbf{Cost and samples.} On the 500-contract run, everything but the uninstrumented original pipeline used about 4.2 million GPT-5.4, 4.5 million GPT-5.4-mini and 0.8 million Gemini 2.5 Flash tokens; grading needs no model calls. Development experiments include the contracts whose earlier stages produced output (93 to 97 of 100), minus one with no scorable contract-specific step, as does the scale run (499 of 500).

\textbf{Licensing, ethics and reproducibility.} Sources keep the FSM-SCG license; no personal data is involved and no contract was deployed publicly. Every number can be recomputed from cached outputs, and the repository linked in the introduction maps each table and figure to its script.

\end{document}